\pdfoutput=1

\documentclass[11pt]{article}

\usepackage[final]{ACL2023}

\usepackage{times}
\usepackage{latexsym}

\usepackage[T1]{fontenc}

\usepackage[utf8]{inputenc}

\usepackage{microtype}

\usepackage{inconsolata}

\usepackage{amsmath}
\usepackage{amssymb}
\usepackage{tikz}
\usepackage{mathabx}
\usepackage{subcaption}
\usepackage[titlenumbered,boxed,ruled]{algorithm2e} 
\makeatletter
\usepackage[final]{pdfpages}
\usepackage{hyperref}
\usepackage{booktabs}       
\usepackage{amsfonts}       
\usepackage{nicefrac}       
\usepackage{microtype}      
\usepackage{xcolor}         
\usepackage[symbol]{footmisc}

\title{ECOLA: Enhancing Temporal Knowledge Embeddings with Contextualized Language Representations}

\author{Zhen Han$^{*1}$, \; Ruotong Liao\thanks{\;\;equal contribution}$^{\;\;1}$, \; Jindong Gu$^{2}$, \; Yao Zhang$^{1}$, \; Zifeng Ding$^{1}$, \; \\ \textbf{Yujia Gu}$^{4}$, \; \textbf{Heinz Köppl}$^{3}$\textbf{,} \; \textbf{Hinrich Schütze}$^{1}$\textbf{,} \; \textbf{Volker Tresp}$^{1}$ \\
$^{1}$Institute of Informatics, LMU Munich $\;$  $^{2}$ Department of Engineering Science, University of Oxford $\;$ \\ $^{3}$Department of Informatics, Technical University of Darmstadt \\ $^{4}$Institute of Informatics, Technical University of Munich\\ 
$\;$ $^{5}$Munich Center for Machine Learning (MCML), Munich, Germany\\
\texttt{hanzhen02111@gmail.com, \; liao@dbs.ifi.lmu.de, \; volker.tresp@lmu.de}\\}

\begin{document}
\maketitle
\begin{abstract}
Since conventional knowledge embedding models cannot take full advantage of the abundant textual information, there have been 
extensive research efforts in enhancing 
 knowledge embedding using texts. 
However, existing enhancement approaches cannot apply to \textit{temporal knowledge graphs} (tKGs), which contain time-dependent event knowledge with complex temporal dynamics. Specifically, existing enhancement approaches often assume knowledge embedding is time-independent. In contrast, the entity embedding in tKG models usually evolves, which poses the challenge of aligning \textit{temporally relevant} texts with entities. To this end, we propose to 
study enhancing temporal knowledge embedding with textual data in this paper. As an approach to this task, we propose \textbf{E}nhanced Temporal Knowledge Embeddings with \textbf{Co}ntextualized \textbf{La}nguage Representations (ECOLA), which takes the temporal aspect into account and injects textual information into temporal knowledge embedding. To evaluate ECOLA, we introduce three new datasets for training and evaluating ECOLA. Extensive experiments show that ECOLA significantly enhances temporal KG embedding models with up to \textbf{287\%} relative improvements regarding Hits@1 on the link prediction task. The code and models are publicly available\footnote{https://anonymous.4open.science/r/ECOLA}.
\end{abstract}

\section{Introduction}

Knowledge graphs (KGs) have long been considered an effective and efficient way to store structural knowledge about the world.  A knowledge graph consists of a collection of \textit{triples} $(s, p, o)$, where $s$ (subject entity) and $o$ (object entity) correspond to nodes, and $p$ (predicate) indicates the edge type (relation) between the two entities. 
Common knowledge graphs \citep{toutanova2015representing, dettmers2018convolutional} assume that the relations between entities are static connections. However, in the real world, there are not only static facts but also time-evolving relations associated with the entities. For example, the political relationship between two countries might worsen  because of trade fights. To this end, temporal knowledge graphs (tKGs) \citep{tresp2015learning} were introduced that capture temporal aspects of relations by extending a triple to a \textit{quadruple}, which adds a timestamp to describe when the relation is valid, e.g. (\textit{Argentina}, \textit{deep comprehensive strategic partnership with}, \textit{China}, \textit{2022 Nov.}).

Conventional knowledge embedding approaches learn KGs by capturing the structure information, suffering from the sparseness of KGs. 
To address this problem, some recent studies incorporate textual information to enrich knowledge embedding. KG-Bert \citep{yao2019kg} takes entity and relation descriptions of a triple as the input of a pre-trained language model (PLM) and turns KG link prediction into a sequence classification problem.  Similarly, KEPLER \citep{wang2021kepler} computes entity representations by encoding entity descriptions with a PLM and then applies KG score functions for link prediction. 
However, they could not be applied to tKGs. 
Specifically, existing approaches (e.g. KEPLER) encode an entity, no matter at which timestamp, with the same static embedding based on a shared entity description. 
In comparison, entity embedding in tKG models usually evolves over time as entities often involve in different events at different timestamps. 
Therefore, we need to use different textual knowledge to augment entity embedding at different timestamps. And it should be taken into account which textual knowledge is relevant to which entity at which timestamp. We name \textbf{this challenge} as \textit{temporal alignment} between texts and tKG, which is to establish a correspondence between textual knowledge and their tKG depiction. 
\textbf{Another challenge} is that many temporal knowledge embedding models \citep{goel2020diachronic, han2020dyernie} learn the entity representations as a function of time. However, the existing enhancement approaches
cannot be naturally applicable to such tKG embedding. 
\begin{figure}
  \centering
  \includegraphics[width=\linewidth]{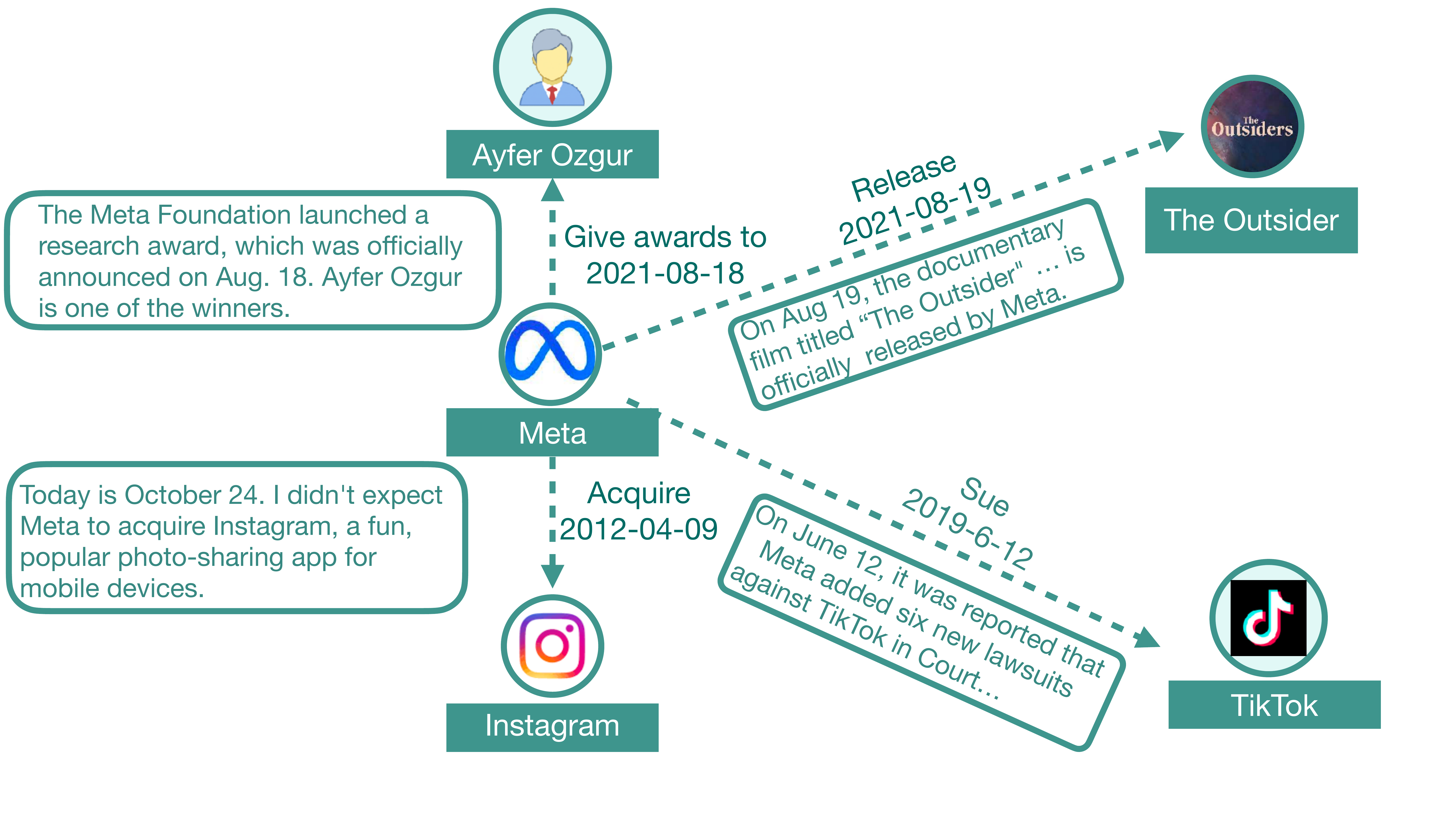}
\caption{\label{fig:ECOLA example} An example of a temporal knowledge graph with textual event descriptions. 
}
\end{figure}
In this work, we propose to study \textit{enhancing temporal knowledge embedding with textual data}. As an approach to this task, we develop \textbf{E}nhanced Temporal Knowledge Embeddings with \textbf{Co}ntextualized \textbf{La}nguage Representations (ECOLA), which uses temporally relevant textual knowledge to enhance the time-dependent knowledge graph embedding. Specifically, we solve the \textbf{temporal alignment challenge} using tKG quadruples as an implicit measure. We pair a quadruple with its relevant textual data, e.g., event descriptions, which corresponds to the temporal relations between entities at a specific time. Then we use the event description to enhance the representations of entities and the predicate involved in the given quadruple. Besides, our approach solves \textbf{the other challenge} using a novel knowledge-text prediction (KTP) task to inject textual knowledge into temporal knowledge embeddings.
Specifically,  
given a quadruple-text pair, we feed PLM with textual tokens concatenated with knowledge embeddings from the tKG model to preserve the original representation capability of temporal knowledge embedding.
The KTP task is an extended masked language modeling task that randomly masks words in texts and entities/predicates/timestamp in quadruples. With the help of the KTP task, ECOLA would be able to recognize mentions of the subject entity and the object entity and align semantic relationships in the text with the predicate in the quadruple. 

For training ECOLA, we need datasets with tKG quadruples and aligned textual event descriptions, which are unavailable in the existing temporal KG benchmarks. 
Thus, we construct three new temporal knowledge graph datasets by adapting two existing datasets, i.e., GDELT \citep{leetaru2013gdelt} and Wiki \citep{dasgupta2018hyte}, and an event extraction dataset \citep{li2020duee}. 

To summarize, our contributions are as follows: (i) We are the first to address the challenge of enhancing temporal knowledge embedding with temporally relevant textual information while preserving the time-evolving properties of entity embedding. 
(ii)  We construct three datasets to train the text-enhanced tKG models. Specifically, we adapt three existing temporal KG completion datasets by augmenting each quadruple with a relevant textual description. 
(iii) Extensive experiments show that ECOLA is model-agnostic and can be potentially combined with any temporal KG embedding model. ECOLA also has a superior performance on the temporal KG completion task and enhances temporal KG 
 models with up to \textbf{287\%} relative improvements in the Hits@1 metric.
(iv) As a joint model, ECOLA also empowers PLMs by integrating \textit{temporal structured knowledge} into them. We select temporal question answering as a downstream NLP task, demonstrating ECOLA's benefit on PLMs.

\section{Preliminaries and Related Work}

\paragraph{Temporal Knowledge Graphs}
Temporal knowledge graphs are multi-relational, directed graphs with labeled timestamped edges between entities (nodes). Let $\mathcal E$ and $\mathcal P$ represent a finite set of entities and predicates, respectively. 
A quadruple $q = (e_s, p, e_o, t)$ represents a timestamped and labeled edge between a subject entity $e_s \in \mathcal E$ and an object entity $e_o \in \mathcal E$ at a timestamp $t \in \mathcal T$. Let $\mathcal F$ represent the set of all true quadruples, 
the temporal knowledge graph completion (tKGC) is the task of inferring $\mathcal F$ based on a set of
observed facts $\mathcal O$. 
Specifically, tKGC is to predict either a missing subject entity $(?, p, e_o, t)$ given the other three components or a missing object entity $(e_s, p, ?, t)$. 
We provide related works on temporal knowledge representations in Appendix \ref{app: TKE}.

 \paragraph{Joint Language and Knowledge Models}
Recent studies have achieved great success in jointly learning language and knowledge representations. 
\citet{zhang2019ernie} and \citet{peters2019knowledge} focus on enhancing language models using external knowledge. They separately pre-train the entity embedding with  knowledge embedding models, e.g., TransE \cite{bordes2013translating}, and inject the pre-trained entity embedding into PLMs, while fixing the entity embedding during training PLMs. Thus, they are not real joint models for learning knowledge embedding and language embedding simultaneously. 
\citet{yao2019kg}, \citet{kim2020multi}, and \citet{wang2021kepler} learn to generate entity embeddings with PLMs from entity descriptions. Moreover, \citet{he2019integrating},  \citet{sun2020colake}, and \citet{liu2020k} exploit the potential of contextualized knowledge representation by constructing subgraphs of structured knowledge and textual data instead of treating single triples as training units. 
Nevertheless, none of these works consider the temporal aspect of knowledge graphs, which makes them different from our proposed ECOLA.

\section{ECOLA}
\begin{figure*}
  \centering
  \includegraphics[width=0.9\linewidth]{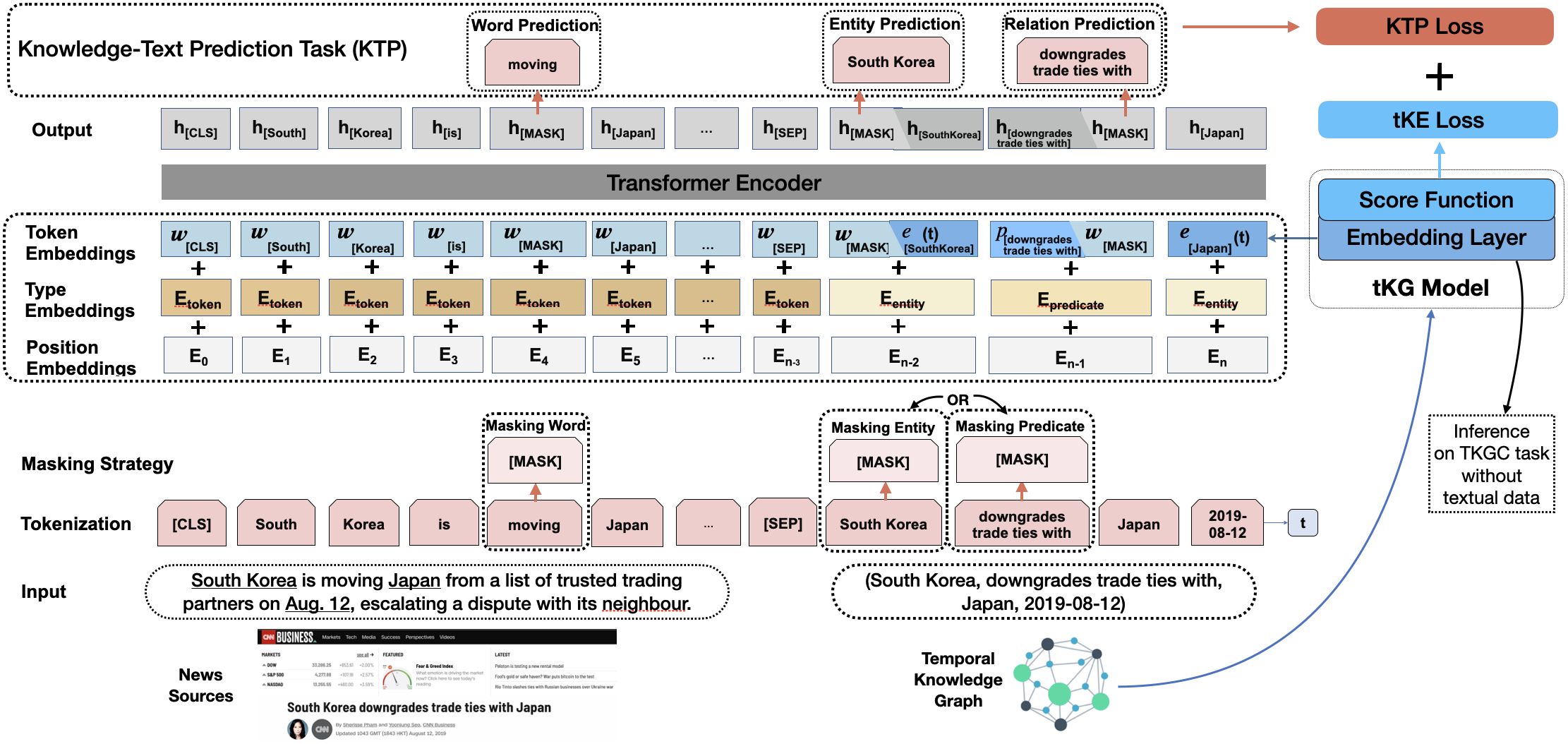}
\caption{\label{fig:ECOLA architecture} Model architecture. ECOLA jointly optimizes the knowledge-text prediction (KTP) objective and the temporal knowledge embedding (tKE) objective. 
}
\end{figure*}

In this section, we present the overall framework
of ECOLA, including the model architecture in Section \ref{sec: embedding layer} - \ref{sec: masked transformer encoder}, a novel task designed for aligning knowledge embedding and language representation in Section \ref{sec: ktp task}, and the training procedure in Section \ref{sec: training procedure}. As shown in Figure \ref{fig:ECOLA architecture}, ECOLA implicitly incorporates textual knowledge into temporal knowledge embeddings by jointly optimizing the \textit{knowledge-text prediction loss} and the \textit{temporal knowledge embedding loss}. Note that, at inference time, we only take the enhanced temporal knowledge embeddings to perform the temporal KG completion task without using PLM and any textual data for preventing information leakage and keeping a fast inference speed.

\subsection{Embedding Layer}
\label{sec: embedding layer}
In tKG embedding models, entity representations evolve over time. Thus, the key point of enhancing a time-dependent entity representation $\mathbf e_i(t)$ is to find texts that are relevant to the entity at the time of interest $t$. 
To this end, we use tKG quadruples (e.g., $(e_i, p, e_j, t)$) 
as an implicit measure for the alignment. We pair a quadruple with its relevant textual data 
and use such textual data to enhance the entity representation $\mathbf e_i(t)$.
Therefore, a training sample is a pair of quadruple from temporal KGs and the corresponding textual description, which are packed together into a sequence. As shown in Figure \ref{fig:ECOLA architecture}, the input embedding is the sum of token embedding, type embedding, and position embedding. For token embedding, we maintain \textbf{three lookup tables} for subwords, entities, and predicates, respectively. 
For subword embedding, we first tokenize the textual description into a sequence of subwords following \citep{devlin2018bert} and use the WordPiece algorithm \citep{wu2016google}. 
As the light blue tokens shown in Figure \ref{fig:ECOLA architecture}, we denote an embedding sequence of subword tokens as $\{\mathbf w_1, ..., \mathbf w_n\}$. In contrast to subword embedding, the embeddings for entities and predicates are directly learned from scratch, similar to common knowledge embedding methods. 
We denote the entity embedding and predicate embedding as $\mathbf e$ and $\mathbf p$, respectively, as the dark blue tokens shown in Figure \ref{fig:ECOLA architecture}.
We separate the knowledge tokens, i.e., entities and predicates, and subword tokens with a special token [SEP]. 
To handle different token types, we add type embedding to indicate the type of each token, i.e., subword, entity, and predicate. For position embedding, we assign each token an index according to its position in the input sequence and follow \citet{devlin2018bert} to apply fully-learnable absolute position embeddings. 

\subsection{Temporal Knowledge Encoder}
\label{sec: temporal knowledge embedding}
As shown in Figure \ref{fig:ECOLA architecture}, the input embedding for \textit{entities} and \textit{predicates} consists of knowledge token embedding, type embedding, and position embedding. In this section, we provide details of the temporal knowledge embedding (tKE) objective. 

A temporal embedding function defines entity embedding as a function that takes an entity and a timestamp as input and generates a time-dependent representation. There is a line of work exploring temporal embedding functions. Since we aim to propose a model-agnostic approach, we combine ECOLA with three temporal embedding functions, i.e., DyERNIE-Euclid \citep{han2020dyernie}, UTEE \citep{han-etal-2021-time}, and DE-SimplE \citep{goel2020diachronic}. In the following, we refer to DyERNIE-Euclid as DyERNIE and take it as an example to introduce our framework. Specifically, the entity representation is derived from an initial embedding and a velocity vector $ \mathbf e_i^{DyER}(t) = \bar{\mathbf e}_i^{DyER} + \mathbf v_{e_i}t$, 
where $\bar{\mathbf e}_i^{DyER}$ represents the initial embedding that does not change over time, and $\mathbf v_{e_i}$ is an entity-specific velocity vector. The combination with other temporal embedding functions is discussed in Section \ref{app: model variants}. 
The score function measuring the plausibility of a quadruple is defined as follows,
\begin{equation}
\label{equa: score function DyERNIE}
\begin{aligned}
& \phi^{DyER}(e_i, p, e_j, t) = \\
& - d(\mathbf P \odot \mathbf e_i^{DyER}(t), \mathbf e_j^{DyER}(t) + \mathbf p ) + b_i + b_j, 
\end{aligned}
\end{equation}
where $\mathbf P$ and $\mathbf p$ represent the predicate matrix and the translation vector of predicate $p$, respectively; $d$ denotes the Euclidean distance, and $b_i$, $b_j$ are scalar biases. By learning tKE, 
we generate $M$ negative samples for each positive quadruple in a batch.  We choose the binary cross entropy as the temporal knowledge embedding objective
\begin{equation}
\mathcal L_{tKE} = \frac{-1}{N} \sum_{k=1}^{N} (y_k 
\log(p_k) + (1 - y_k)\log(1 - p_k)),
\end{equation}
where $N$ is the sum of positive and negative training samples, $y_k$ represents the binary label indicating whether a training sample is positive or not, $p_k$ denotes the predicted probability $\sigma(\phi^{DyER}_k)$, and $\sigma(\cdot)$ represents the sigmoid function.

\subsection{Masked Transformer Encoder}
\label{sec: masked transformer encoder}
To encode the input sequence, we use the pre-trained language representation model Bert \citep{devlin2018bert}. 
Specifically, the encoder feeds a sequence of $N$ tokens including \textit{entities}, \textit{predicates}, and \textit{subwords} into the embedding layer introduced in Section \ref{sec: embedding layer} to get the input embeddings and then computes $L$ layers of $d$-dimensional contextualized representations. 
Eventually, we get a contextualized representation for each token, which could be further used to predict masked tokens.

\subsection{Knowledge-Text Prediction Task}
\label{sec: ktp task}
To incorporate textual knowledge into temporal knowledge embedding, we use the pre-trained language model Bert to encode the textual description and propose a knowledge-text prediction task to align the language representations and the knowledge embedding. The knowledge-text prediction task is an extension of the masked language modeling (MLM) task. 
As illustrated in Figure \ref{fig:ECOLA architecture}, given a 
pair of a quadruple and the corresponding event description, the knowledge-text prediction task is to randomly mask some of the input tokens and train the model to predict the original index of the masked tokens based on their contexts. As different types of tokens are masked, we encourage ECOLA to learn different capabilities:

\begin{itemize}
\item \textbf{Masking entities}. 
To predict an entity token in the quadruple, ECOLA has the following ways to gather information. First, the model can detect the textual mention of this entity token and determine the entity; second, if the other entity token and the predicate token are not masked, the model can utilize the available knowledge token to make a prediction, which is similar to the traditional semantic matching-based temporal KG models. Masking entity nodes helps ECOLA align the representation spaces of language and structured knowledge, and inject contextualized representations into entity embeddings.

\item \textbf{Masking predicates}.
To predict the predicate token in the quadruple, the model needs to detect mentions of the subject entity and object entity and classify the semantic relationship between the two entity mentions. Thus, masking predicate tokens helps the model integrate language representation into the predicate embedding and map words and entities into a common representation space.

\item \textbf{Masking subwords}.
When subwords are masked, the objective is similar to traditional MLM. The difference is that ECOLA considers not only the dependency information in the text but also the entities and the logical relationship in the quadruple. Additionally, we initialize the encoder with the pre-trained BERT$_{\textrm{base}}$. Thus, masking subwords helps ECOLA keep linguistic knowledge and avoid catastrophic forgetting while integrating contextualized representations into temporal knowledge embeddings.
\end{itemize}

In each quadruple, the predicate and each entity have a probability of 15\% to be masked. Similarly, we mask 15\% of the subwords of the textual description at random. We ensure that entities and the predicate cannot be masked at the same time in a single training sample, where we conduct an ablation study in Section \ref{sec: experiments} to show the improvement of making this constraint. When a token is masked, we replace it with (1) the [MASK] token 80\% of the time, (2) a randomly sampled token with the same type as the original token 10\% of the time, (3) the unchanged token 10\% of the time. For each masked token, the contextualized representation in the last layer of the encoder is used for three classification heads, which are responsible for predicting entities, predicates, and subword tokens, respectively. At last, a cross-entropy loss $\mathcal L_{KTP}$ is calculated over these masked tokens.

\subsection{Training Procedure and Inference}
\label{sec: training procedure}
We initialize the transformer encoder with BERT$_{\textrm{base}}$\footnote{https://huggingface.co/bert-base-uncased} and the knowledge encoder with random vectors. Then we use the temporal knowledge embedding (tKE) objective $\mathcal L_{tKE}$ to train the knowledge encoder and use the knowledge-text prediction (KTP) objective $\mathcal L_{KTP}$ to incorporate temporal factual knowledge and textual knowledge in the form of a multi-task loss:
\begin{equation*}
\mathcal L = \mathcal L_{tKE} + \lambda \mathcal L_{KTP},
\end{equation*}
where $\lambda$ is a hyperparameter to balance tKE loss and KTP loss. Note that those two tasks share the same embedding layer of entities and predicates. 
At inference time,  we aim to answer link prediction queries, e.g., $(e_s, p, ?, t)$. Since there is no textual description at inference time, we take the entity and predicate embedding as input and use the score function of the knowledge encoder, e.g., Equation \ref{equa: score function DyERNIE}, to predict the missing links. Specifically, the score function assigns a plausibility score to each quadruple, and the proper object can be inferred by ranking the scores of all quadruples $\{(e_s, p, e_j, t), e_j \in \mathcal E\}$ that are accompanied with candidate entities.

\section{Model-Agnostics}
\label{app: model variants}
ECOLA is model-agnostic and can enhance different temporal knowledge embedding models. Besides ECOLA-DyERNIE, we introduce here two additional variants of ECOLA.

\paragraph{ECOLA-DE} enhances DE-SimplE, which applies the diachronic embedding (DE) function \citep{goel2020diachronic}. DE-function defines the temporal embeddings of entity $e_i$ at timestamp $t$ as 
\begin{equation}
\mathbf e_i^{DE}(t)[n] = \left\{  
           \begin{aligned}
            &\mathbf a_{e_i}[n]  \;\;\;\; \text{if} \;\; 1 \le n \le \gamma d, \\
		  &\mathbf a_{e_i}[n] \sin (\boldsymbol \omega_{e_i}[n]t + \mathbf b_{e_i}[n]) \; \text{else.} \\
\end{aligned}
\right.   
\label{equa: de-encoding}
\end{equation}
Here, $\mathbf e_i^{DE}(t)[n]$ denotes the $n^{th}$ element of the embeddings of entity $e_i$ at time $t$. $\mathbf a_{e_i}, \boldsymbol \omega_{e_i}, \mathbf b_{e_i} \in \mathbb R^d$ are entity-specific vectors with learnable parameters, $d$ is the dimensionality, and $\gamma \in [0, 1]$ represents the portions of the time-independent part. 


\paragraph{ECOLA-UTEE} enhances UTEE \cite{han-etal-2021-time} that learns a \textit{shared} temporal encoding for all entities to address the overfitting problem of DE-SimplE on sparse datasets. Compared to ECOLA-DE, ECOLA-UTEE replaces Equation \ref{equa: de-encoding} with $\mathbf e_i^{UTEE}(t) = [ \bar{\mathbf e_i} || \mathbf a \sin (\boldsymbol \omega t + \mathbf b)], \bar{\mathbf e_i} \in \mathbb R^{\gamma d}; \mathbf a, \mathbf w, \mathbf b \in \mathbb R^{(1-\gamma)d}$,
where $\bar{\mathbf e}_i$ denotes entity-specific time-invariant part,  $||$ denotes concatenation, $\mathbf a$, $\boldsymbol \omega$, and $\mathbf b$ are shared among all entities.

\begin{table*}[htbp]
    \caption{Temporal link prediction results: Mean Reciprocal Rank (MRR, \%) and Hits@1/3(\%). The results of the proposed fusion models (in bold) and their counterpart KG models are listed together. 
    }
    \label{tab: link prediction results PART1}
    \begin{center}
      \resizebox{.7\textwidth}{!}{
    \begin{tabular}{l|ccc|ccc|ccc} 
      \toprule 
     \multicolumn{1}{l}{Datasets} & \multicolumn{3}{|c}{\textbf{GDELT - filtered}} &  \multicolumn{3}{|c}{\textbf{Wiki - filtered}} & \multicolumn{3}{|c}{\textbf{DuEE - filtered}}\\
      \midrule 
      Model & MRR & Hits@1 & Hits@3 & MRR & Hits@1 & Hits@3  & MRR & Hits@1 & Hits@3 \\
       \midrule 
        TransE &  8.08&  0.00 & 8.33 & 27.25 & 16.09 & 33.06 &  34.25 & 4.45 & 60.73\\
        SimplE & 10.98 & 4.76 & 10.49 & 20.75 & 16.77 & 23.23 & 51.13 & 40.69 & 58.30\\
        DistMult & 11.27 & 4.86 & 10.87 & 21.40 & 17.54 & 23.86 & 48.58 & 38.26 & 55.26\\
      \midrule 
      TeRO & 6.59 & 1.75 & 5.86 & 32.92 & 21.74 & 39.12 & 54.29 & 39.27 & 63.16\\
      ATiSE  & 7.00 & 2.48 & 6.26 & 35.36 & 24.07 & 41.69 & 53.79 & 42.31 & 59.92\\
       TNTComplEx & 8.93 & 3.60 & 8.52 & 34.36 & 22.38 & 40.64 & 57.56 & 43.52 & 65.99\\
       \midrule 
       UTEE & 9.76 & 4.23 & 9.77 & 26.96 & 20.98 & 30.39 & 53.36 & 43.92 & 60.52 \\
       \textbf{ECOLA-UTEE} & 19.11 $\pm$ & 15.29 $\pm$ & 19.46 $\pm$ & 38.35 $\pm$ & 30.56 $\pm$ & 42.11 $\pm$ & \textbf{60.36} $\pm$ & \textbf{46.55} $\pm$ & 69.22 $\pm$\\
          & 00.16 & 00.38 & 00.05 & 00.22 & 00.18 & 00.14 & 00.36 & 00.51 & 00.93\\
      \midrule 
      DyERNIE & 10.72 & 4.24 & 10.81 & 23.51 & 14.53 & 25.21 & 57.58 & 41.49 & \textbf{70.24}\\
      \textbf{ECOLA-DyERNIE} & \textbf{19.99} $\pm$ & \textbf{16.40} $\pm$ & \textbf{19.78} $\pm$ & \textbf{41.22} $\pm$ & \textbf{33.02} $\pm$ & \textbf{45.00} $\pm$  & 59.64 $\pm$ & 46.35 $\pm$ & 67.87 $\pm$\\
        & 00.05 & 00.09 & 00.03 & 00.04 & 00.06 & 00.27 & 00.20 & 00.18 & 00.53\\
      \bottomrule 
    \end{tabular} }
    \end{center}
	\end{table*}

\section{Datasets}
\label{sec: datasets}
Training ECOLA requires both temporal KGs and textual descriptions. Given a quadruple $(e_s, p, e_o, t)$, the key point is to find texts that are temporally relevant to $e_s$ and $e_o$ at $t$. Existing tKG datasets do not provide such information. To facilitate the research on integrating textual knowledge into temporal knowledge embedding, we reformat GDELT\footnote{https://www.gdeltproject.org/data.html\#googlebigquery}, DuEE\footnote{https://ai.baidu.com/broad/download}, and Wiki\footnote{https://www.wikidata.org/wiki/Wikidata:Main\_Page}. We show the dataset statistics in Table \ref{tab:dataset statistics} in the appendix. 

\paragraph{GDELT} is an initiative knowledge base storing events across the globe connecting people and organizations, e.g., (\textit{Google, consult, the United States, 2018/01/06}). 
For each quadruple, GDELT provides the link to the news report which the quadruple is extracted from.  
We assume each sentence that contains both mentions of the subject and object is relevant to the given quadruple, and thus, temporally aligned with the subject and object at the given timestamp. We pair each of these sentences with the given quadruple to form a training sample. This process is similar to the distant supervision algorithm \cite{mintz2009distant} in the relation extraction task. 
The proposed dataset contains 5849 entities, 237 predicates, 2403 timestamps, and 943956 quadruples with accompanying sentences. 

\paragraph{DuEE} is originally a human-annotated dataset for event extraction containing 65 event types and 121 argument roles. Each sample contains a sentence and several extracted event tuples. We reformat DuEE by manually converting event tuples into quadruples and then pairing quadruples with their corresponding sentence.

\paragraph{Wiki} is a temporal KG dataset proposed by \citet{leblay2018deriving}. 
 Following the post-processing by \citet{dasgupta2018hyte}, we discretize the time span into 82 different timestamps.   
We align each entity to its Wikipedia page and extract the first section as its description. To construct the relevant textual data of each quadruple, we combine the subject description, relation, and object description into a sequence. 
In this case, the knowledge-text prediction task lets the subject entity see the descriptions of its \textit{neighbors at different timestamps}, thus, preserving the temporal alignment between time-dependent entity representation and textual data.


\section{Experiments}
\label{sec: experiments}
\begin{figure}
\begin{subfigure}{0.54\linewidth}
 \centering
   \includegraphics[width=\linewidth]{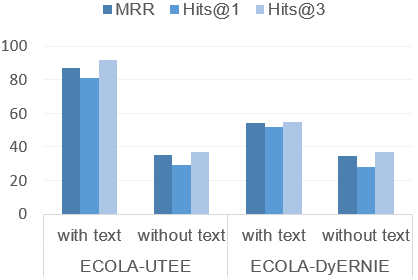}
   \caption{\label{fig: inf_w_text}}
\end{subfigure}%
\begin{subfigure}{0.45\linewidth}
 \centering
   \includegraphics[width=\linewidth]{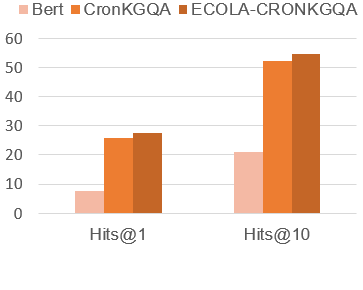}
   \caption{\label{fig: tkgqa}}
\end{subfigure}%
 \caption{ \label{fig: with/without texts and TQA results.} (a) Results of tKG completion task on GDELT with and without using textual description of test quadruples. (b) ECOLA benefits language representations on the temporal question answering task. 
 } 
\end{figure}

We evaluate the enhanced temporal knowledge embedding on the temporal KG completion task. Specifically, we take the entity and predicate embedding of ECOLA-DyERNIE and use Equation \ref{equa: score function DyERNIE} to predict missing links. 
The textual description of test quadruples could introduce essential information and make the completion task much easier. Thus, to make a \textbf{fair comparison} with other temporal KG embedding models, we take the enhanced \textit{lookup table embedding} of temporal KGs to perform the link prediction task at test time but use neither textual descriptions of test quadruples nor the language model. We report such results in Table \ref{tab: link prediction results PART1}.
As additional results, we also show the prediction outcome that takes the text description of test quadruples as input in Figure \ref{fig: inf_w_text}.

\paragraph{Baselines} We include both static and temporal KG embedding models. From the static KG embedding models, we use TransE \citep{bordes2013translating}, DistMult \citep{yang2014embedding}, and SimplE \citep{kazemi2018simple}. These methods ignore the time information. From the temporal KG embedding models, we compare our model with several state-of-the-art methods, including AiTSEE \citep{xu2019temporal}, TNTComplE\citep{lacroix2020tensor}, DyERNIE\footnote{For a fair comparison with other baselines, we choose DyERNIE-Euclid.} \citep{han2020dyernie}, TeRO \citep{xu-etal-2020-tero}, and DE-SimplE \citep{goel2020diachronic}. 
We provide implementation details in Appendix \ref{app: experimental setting} and attach the source code in the supplementary material.

\paragraph{Evaluation Protocol}
For each quadruple $q = (e_s, p, e_o, t)$ in the test set $\mathcal G_{test}$, we create two queries: $(e_s, p, ?, t)$ and  $(?, p, e_o, t)$. For each query, the model ranks all possible entities $\mathcal E$ according to their scores. 
Let $\textit{Rank}({e_s})$ and $\textit{Rank}({e_o})$ represent the rank for $e_s$ and $e_o$ of the two queries respectively, we evaluate our models using standard metrics across the link prediction literature: \textit{mean reciprocal rank (MRR)}: $\frac{1}{2\cdot |\mathcal G_{test}|} \sum_{q \in \mathcal G_{test}}(\frac{1}{\textit{Rank}({e_s})} + \frac{1}{\textit{Rank}({e_o})})$ and \textit{Hits}$@k(k \in \{1,3,10\})$: the percentage of times that the true entity candidate appears in the top $k$ of ranked candidates.

\paragraph{Quantitative Study}
Table \ref{tab: link prediction results PART1} reports the tKG completion results on the test sets, which are averaged over three trials. 
Firstly, we can see that ECOLA-UTEE improves its baseline temporal KG embedding model, UTEE, by a large margin, demonstrating the effectiveness of our fusing strategy. Specifically, ECOLA-UTEE enhances UTEE on GDELT with a \textit{relative improvement} of \textbf{95\%} and \textbf{99\%} in terms of mean reciprocal rank (MRR) and Hits@3, even nearly \textbf{four times} better in terms of Hits@1. Thus, its superiority is clear on GDELT, which is the most challenging dataset among benchmark tKG datasets, containing nearly one million quadruples. Secondly, ECOLA-UTEE and ECOLA-DE generally outperform UTEE and DE-SimplE on the three datasets, demonstrating that ECOLA is model-agnostic and can enhance different tKG embedding models. Besides, in the DuEE dataset, ECOLA-DyERNIE achieves a better performance than DyERNIE in Hits@1 and MRR, but the gap reverses in Hits@3. The reason could be that ECOLA-DyERNIE is good at classifying hard negatives using textual knowledge, and thus has a high Hits@1; however, since DuEE is much smaller than the other two datasets, ECOLA-DyERNIE may overfit in some cases, where the ground truth is pushed away from the top 3 ranks. 

\begin{figure}
\begin{subfigure}{0.5\linewidth}
 \centering
   \includegraphics[width=\linewidth]{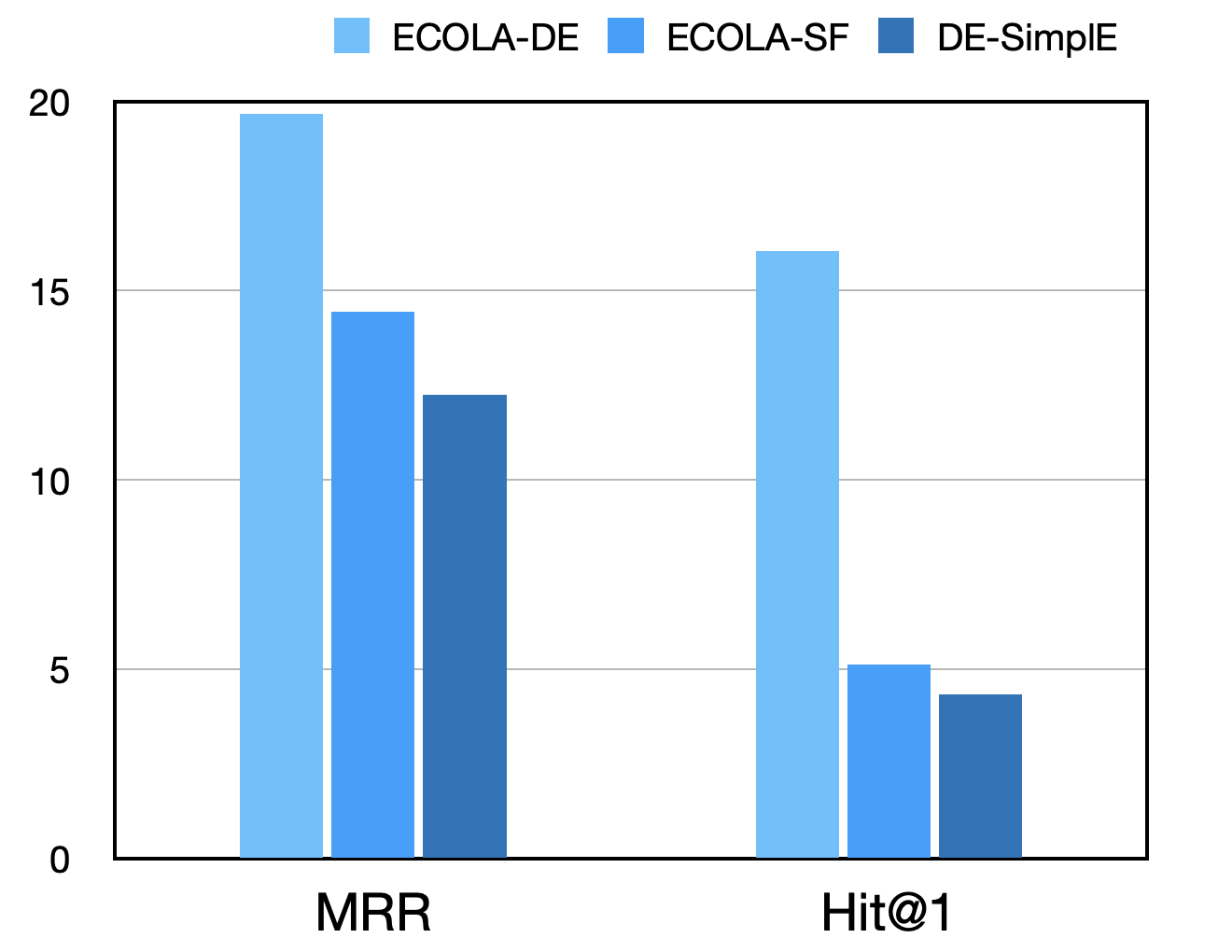}
   \caption{\label{fig: ablation study ECOLA-SF vs. ECOLA-DE vs. DE}}
\end{subfigure}%
\begin{subfigure}{0.5\linewidth}
 \centering
   \includegraphics[width=\linewidth]{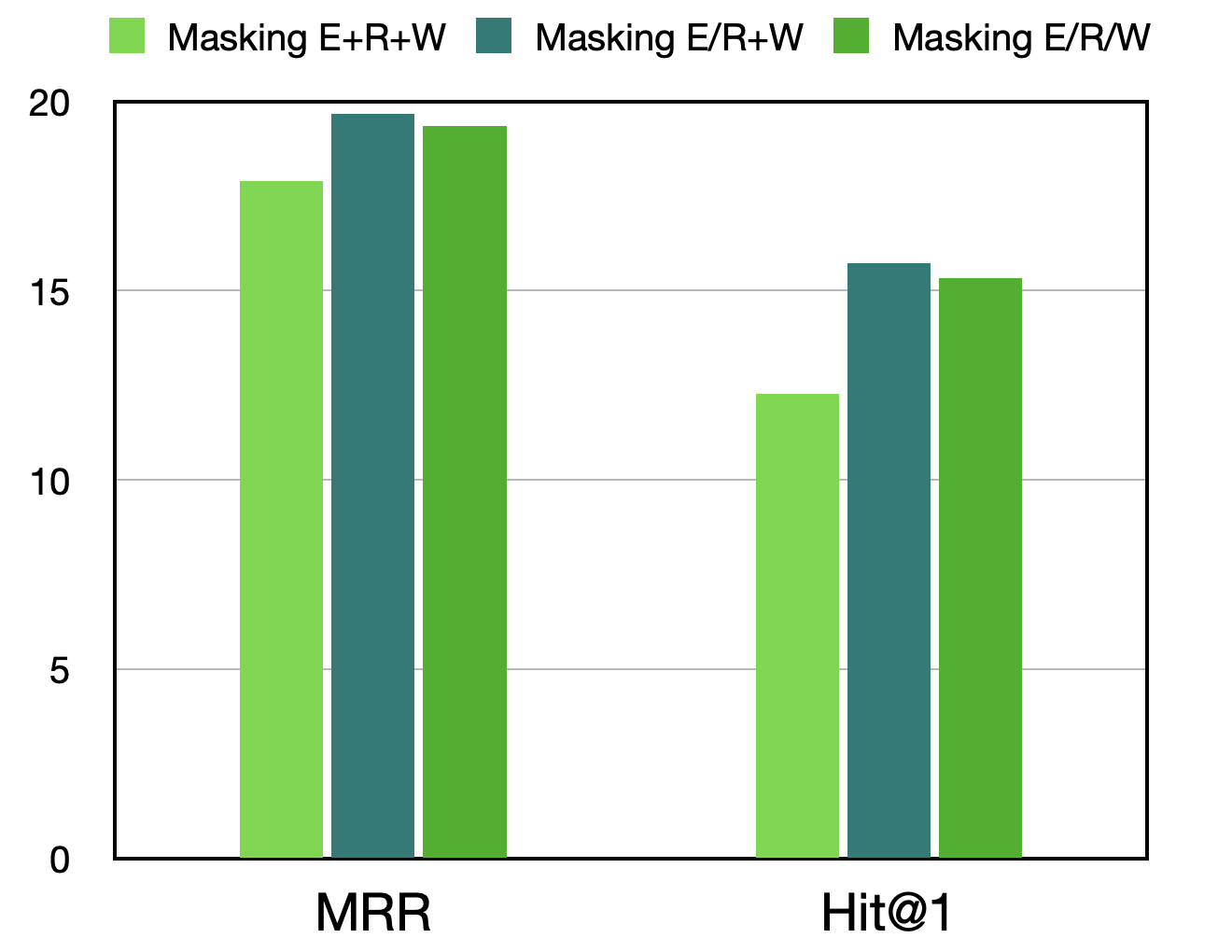}
   \caption{\label{fig: ablation study Masking Strategy}}
\end{subfigure}%
 \caption{ \label{fig:Ablation Study.} Ablation Study. (a) Temporal alignment analysis. We compare De-SimplE, ECOLA-DE, and ECOLA-SF in terms of MRR(\%) and Hits@1(\%) on GDELT. (b) Masking strategy analysis. We compare ECOLA-DE with different masking strategies and show the results of MRR(\%) and Hits@1(\%) on GDELT. 
 } 
\end{figure}

\paragraph{Ablation Study}
We compare DE-SimplE, ECOLA-DE, and ECOLA-SF on GDELT in Figure \ref{fig: ablation study ECOLA-SF vs. ECOLA-DE vs. DE}. 
ECOLA-SF is the \textbf{static counterpart} of ECOLA-DE, where we do not consider the temporal alignment while incorporating textual knowledge. Specifically, ECOLA-SF integrates all textual knowledge into the \textbf{time-invariant part} of entity representations. We provide more details of ECOLA-SF in Appendix \ref{app: ecola-sf}. 
In particular, the performance gap between ECOLA-DE and ECOLA-SF is significant, demonstrating the \textit{temporal alignment} between time-dependent entity representation and textual knowledge is more powerful than the \textit{static alignment}. 
 Moreover, Figure \ref{fig: ablation study Masking Strategy} shows the results of different masking strategies on GDELT. The first strategy, e.g, \textit{Masking E+R+W}, allows to simultaneously mask predicate, entity, and subword tokens in the same training sample. The second strategy is \textit{Masking E/R+W}, where we mask 15\% subword tokens in the language part, \textit{and either} an entity \textit{or} a predicate in the knowledge tuple. 
 In the third strategy called \textit{Masking E/R/W}, for each training sample, we choose to mask \textit{either} subword tokens, an entity, \textit{or} the predicate. Figure \ref{fig: ablation study Masking Strategy} shows the advantage of the second masking strategy, indicating that remaining adequate information in the knowledge tuple helps the model to align the knowledge embedding and language representations.

\paragraph{Qualitative Analysis} 
To investigate why incorporating textual knowledge can improve the tKG embedding models' performance, we study the test samples that have been correctly predicted by the fusion model ECOLA-DE but wrongly by the tKG model DE-SimplE. It is observed that language representations help overcome the incompleteness of the tKG by leveraging knowledge from augmented textual data. For example, there is a test quadruple \textit{(US, host a visit, ?, 19-11-14)} with ground truth \textit{R.T. Erdoğan}. The training set contains a quite relevant quadruple, i.e., \textit{(Turkey, intend to negotiate with, US, 19-11-11)}. However, the given tKG does not contain information indicating that the entity \textit{R.T. Erdoğan} is a representative of \textit{Turkey}. 
So it is difficult for the tKG model DE-SimplE to infer the correct answer from the above-mentioned quadruple. In ECOLA-DE, the augmented textual data do contain such information, e.g. \textit{"The president of Turkey, R.T. Erdogan, inaugurated in Aug. 2014."}, which narrows the gap between \textit{R.T. Erdogan} and \textit{Turkey}. Thus, by integrating textual information into temporal knowledge embedding, the enhanced model can gain additional information which the knowledge base does not include.

\section{Discussion}

\paragraph{Inference with Textual Data}
In Section \ref{sec: experiments} we do fair comparisons with tKG embedding models where textual data of test quadruples is absent during inference time. However, if the textual descriptions of the test quadruples are given during inference, will the contextualized language model incorporate this information into tKG embeddings? We use the entity predictor of the knowledge-text prediction task to perform the tKG completion task on GDELT.
As shown in Figure \ref{fig: inf_w_text}, the results show significant improvement across all metrics, specifically, \textbf{145\%} relatively higher regarding MRR of ECOLA-UTEE when given textual data during inference than not given. Thus, the results confirm that KTP task is a good choice for successful alignment between knowledge and language space and ECOLA utilizes the pre-trained language model to inject language representations into temporal knowledge embeddings.

\paragraph{Masking Temporal Information in KTP}
\label{paragraph: Masking time}
As temporal alignment is crucial for enhancing temporal knowledge embeddings, we study the effect of masking temporal information by extending the existing KTP task with an additional \textit{time prediction task}, where the timestamp in the input is masked, and the model learns to predict the original timestamp. 
The extended model is named as tECOLA-UTEE and has significant performance gain on both GDELT and Wiki datasets across all metrics as shown in Table \ref{tab: tKTP}.
We conjecture that the additional time prediction task forces the model to capture the temporal dynamics in temporal knowledge embeddings and utilize the temporal information in given textual descriptions.
Since each temporal knowledge embedding models the temporal information in different ways, masking and predicting temporal information will be specific to each temporal knowledge embedding model. We leave this finding to future work for further inspections.

\paragraph{Temporal Question Answering}
\label{paragraph: TKGQA results}
Although we focus on generating informative temporal knowledge embeddings in this work, joint models often benefit both the language model and the temporal KG model mutually. 
Besides the tKG completion task, we evaluate the enhanced language model in ECOLA on the temporal question-answering task to study the enhancement of the language model. 
\cite{saxena2021cronkgqa} introduced the dataset \textsc{CronQuestions} containing natural temporal questions with different types of temporal constraints. They proposed a baseline \textsc{CronKGQA} that uses BERT to understand the temporal constraints followed by a scoring function for answer prediction. We enhance \textsc{CronKGQA} with ECOLA and name the enhanced model as ECOLA-\textsc{CronKGQA}. We sample a subset of \textsc{CronQuestions} and build a corresponding text-quadruple paired dataset to train ECOLA-\textsc{CronKGQA}.
Figure \ref{fig: tkgqa} shows that our proposed ECOLA enhances CronKGQA with \textbf{7.4 \% relative improvements} regarding precision, demonstrating the benefits of ECOLA to the language model. 

\begin{table}[htbp]
    \caption{ Performance of masking temporal information on the knowledge-text prediction task.}
    \centering
    \label{tab: tKTP}
      \resizebox{.45\textwidth}{!}{
    \begin{tabular}{l|ccc|ccc}
      \toprule 
     \multicolumn{1}{l}{Datasets} & \multicolumn{3}{|c}{\textbf{GDELT - filtered}} &  \multicolumn{3}{|c}{\textbf{Wiki - filtered}} \\
      \midrule 
      Model & MRR & Hits@1 & Hits@3 & MRR & Hits@1 & Hits@3 \\
       \midrule %
       ECOLA-UTEE & 19.11  & 15.29 & 19.46  & 38.35  & 30.56  & 42.11   \\
       tECOLA-UTEE & 20.39 & 16.83 & 20.08  & 42.53 & 34.06 & 46.32 \\
        & (6.7\% $\uparrow$) & (10.1\% $\uparrow$) & (3.2\% $\uparrow$) & (10.9\% $\uparrow$) & (11.5\% $\uparrow$) & (10.0\% $\uparrow$) \\
      \bottomrule 
    \end{tabular}
     }
	\end{table}

\section{Conclusion}
We propose ECOLA 
to enhance time-evolving entity representations with temporally relevant textual data 
using a novel knowledge-text prediction task. 
Besides, we construct three datasets that contain paired structured temporal knowledge and unstructured textual descriptions, which can benefit future research on fusing \textit{temporal} structured and unstructured knowledge. 
Extensive experiments show ECOLA can improve various temporal knowledge graph models by a large margin. 

\section*{Limitations}
To train ECOLA, we need to provide structured knowledge with aligned unstructured textual data to the model. Thus, we should either manually pair quadruples with event descriptions or use some matching algorithm to automatically build the pairs. The former requires human labeling effort and is hard to apply on large-scale datasets, while the latter would introduce noise into the dataset. Thus, ECOLA is currently tailored for domain adaptation and enhances pre-trained models with domain knowledge. There is still work to be done to let models be jointly trained on \textit{large-scale} structured and unstructured data.


\section*{Ethics Statement}
ECOLA is tailored to integrate temporal knowledge embedding and textual knowledge and can be applied to 
a wide variety of downstream tasks, such as temporal knowledge graph link prediction and temporal question answering. It can also power search and, thus, serve
as a key intermediary of information in users' lives. Since most temporal knowledge graphs are automatically extracted from web data, it's important to ensure it does not contain offensive content. ECOLA can be used to classify the quadruples in temporal knowledge graphs using the pre-trained language model and contribute to the knowledge graph protection's perspective.

\bibliography{custom} 

\begin{thebibliography}{30}
\expandafter\ifx\csname natexlab\endcsname\relax\def\natexlab#1{#1}\fi

\bibitem[{Bordes et~al.(2013)Bordes, Usunier, Garcia-Duran, Weston, and
  Yakhnenko}]{bordes2013translating}
Antoine Bordes, Nicolas Usunier, Alberto Garcia-Duran, Jason Weston, and Oksana
  Yakhnenko. 2013.
\newblock Translating embeddings for modeling multi-relational data.
\newblock \emph{Advances in neural information processing systems}, 26.

\bibitem[{Dasgupta et~al.(2018)Dasgupta, Ray, and Talukdar}]{dasgupta2018hyte}
Shib~Sankar Dasgupta, Swayambhu~Nath Ray, and Partha Talukdar. 2018.
\newblock Hyte: Hyperplane-based temporally aware knowledge graph embedding.
\newblock In \emph{Proceedings of the 2018 conference on empirical methods in
  natural language processing}, pages 2001--2011.

\bibitem[{Dettmers et~al.(2018)Dettmers, Minervini, Stenetorp, and
  Riedel}]{dettmers2018convolutional}
Tim Dettmers, Pasquale Minervini, Pontus Stenetorp, and Sebastian Riedel. 2018.
\newblock Convolutional 2d knowledge graph embeddings.
\newblock In \emph{Proceedings of the AAAI conference on artificial
  intelligence}, volume~32.

\bibitem[{Devlin et~al.(2018)Devlin, Chang, Lee, and
  Toutanova}]{devlin2018bert}
Jacob Devlin, Ming-Wei Chang, Kenton Lee, and Kristina Toutanova. 2018.
\newblock Bert: Pre-training of deep bidirectional transformers for language
  understanding.
\newblock \emph{arXiv preprint arXiv:1810.04805}.

\bibitem[{Goel et~al.(2020)Goel, Kazemi, Brubaker, and
  Poupart}]{goel2020diachronic}
Rishab Goel, Seyed~Mehran Kazemi, Marcus Brubaker, and Pascal Poupart. 2020.
\newblock Diachronic embedding for temporal knowledge graph completion.
\newblock In \emph{Proceedings of the AAAI Conference on Artificial
  Intelligence}, volume~34, pages 3988--3995.

\bibitem[{Han et~al.(2020)Han, Ma, Chen, and Tresp}]{han2020dyernie}
Zhen Han, Yunpu Ma, Peng Chen, and Volker Tresp. 2020.
\newblock Dyernie: Dynamic evolution of riemannian manifold embeddings for
  temporal knowledge graph completion.
\newblock \emph{arXiv preprint arXiv:2011.03984}.

\bibitem[{Han et~al.(2021)Han, Zhang, Ma, and Tresp}]{han-etal-2021-time}
Zhen Han, Gengyuan Zhang, Yunpu Ma, and Volker Tresp. 2021.
\newblock \href {https://doi.org/10.18653/v1/2021.emnlp-main.639}
  {Time-dependent entity embedding is not all you need: A re-evaluation of
  temporal knowledge graph completion models under a unified framework}.
\newblock In \emph{Proceedings of the 2021 Conference on Empirical Methods in
  Natural Language Processing}, pages 8104--8118, Online and Punta Cana,
  Dominican Republic. Association for Computational Linguistics.

\bibitem[{He et~al.(2019)He, Zhou, Xiao, Liu, Yuan, Xu
  et~al.}]{he2019integrating}
Bin He, Di~Zhou, Jinghui Xiao, Qun Liu, Nicholas~Jing Yuan, Tong Xu, et~al.
  2019.
\newblock Integrating graph contextualized knowledge into pre-trained language
  models.
\newblock \emph{arXiv preprint arXiv:1912.00147}.

\bibitem[{Kazemi and Poole(2018)}]{kazemi2018simple}
Seyed~Mehran Kazemi and David Poole. 2018.
\newblock Simple embedding for link prediction in knowledge graphs.
\newblock \emph{Advances in neural information processing systems}, 31.

\bibitem[{Kim et~al.(2020)Kim, Hong, Ko, and Seo}]{kim2020multi}
Bosung Kim, Taesuk Hong, Youngjoong Ko, and Jungyun Seo. 2020.
\newblock Multi-task learning for knowledge graph completion with pre-trained
  language models.
\newblock In \emph{Proceedings of the 28th International Conference on
  Computational Linguistics}, pages 1737--1743.

\bibitem[{Kingma and Ba(2014)}]{kingma2014adam}
Diederik~P Kingma and Jimmy Ba. 2014.
\newblock Adam: A method for stochastic optimization.
\newblock \emph{arXiv preprint arXiv:1412.6980}.

\bibitem[{Lacroix et~al.(2020)Lacroix, Obozinski, and
  Usunier}]{lacroix2020tensor}
Timoth{\'e}e Lacroix, Guillaume Obozinski, and Nicolas Usunier. 2020.
\newblock Tensor decompositions for temporal knowledge base completion.
\newblock \emph{arXiv preprint arXiv:2004.04926}.

\bibitem[{Leblay and Chekol(2018)}]{leblay2018deriving}
Julien Leblay and Melisachew~Wudage Chekol. 2018.
\newblock Deriving validity time in knowledge graph.
\newblock In \emph{Companion Proceedings of the The Web Conference 2018}, pages
  1771--1776.

\bibitem[{Leetaru and Schrodt(2013)}]{leetaru2013gdelt}
Kalev Leetaru and Philip~A Schrodt. 2013.
\newblock Gdelt: Global data on events, location, and tone, 1979--2012.
\newblock In \emph{ISA annual convention}, volume~2, pages 1--49. Citeseer.

\bibitem[{Li et~al.(2020)Li, Li, Pan, Chen, Peng, Wang, Lyu, and
  Zhu}]{li2020duee}
Xinyu Li, Fayuan Li, Lu~Pan, Yuguang Chen, Weihua Peng, Quan Wang, Yajuan Lyu,
  and Yong Zhu. 2020.
\newblock Duee: a large-scale dataset for chinese event extraction in
  real-world scenarios.
\newblock In \emph{CCF International Conference on Natural Language Processing
  and Chinese Computing}, pages 534--545. Springer.

\bibitem[{Liu et~al.(2020)Liu, Zhou, Zhao, Wang, Ju, Deng, and Wang}]{liu2020k}
Weijie Liu, Peng Zhou, Zhe Zhao, Zhiruo Wang, Qi~Ju, Haotang Deng, and Ping
  Wang. 2020.
\newblock K-bert: Enabling language representation with knowledge graph.
\newblock In \emph{Proceedings of the AAAI Conference on Artificial
  Intelligence}, volume~34, pages 2901--2908.

\bibitem[{Ma et~al.(2019)Ma, Tresp, and Daxberger}]{ma2019embedding}
Yunpu Ma, Volker Tresp, and Erik~A Daxberger. 2019.
\newblock Embedding models for episodic knowledge graphs.
\newblock \emph{Journal of Web Semantics}, 59:100490.

\bibitem[{Mintz et~al.(2009)Mintz, Bills, Snow, and
  Jurafsky}]{mintz2009distant}
Mike Mintz, Steven Bills, Rion Snow, and Dan Jurafsky. 2009.
\newblock Distant supervision for relation extraction without labeled data.
\newblock In \emph{Proceedings of the Joint Conference of the 47th Annual
  Meeting of the ACL and the 4th International Joint Conference on Natural
  Language Processing of the AFNLP}, pages 1003--1011.

\bibitem[{Peters et~al.(2019)Peters, Neumann, Logan~IV, Schwartz, Joshi, Singh,
  and Smith}]{peters2019knowledge}
Matthew~E Peters, Mark Neumann, Robert~L Logan~IV, Roy Schwartz, Vidur Joshi,
  Sameer Singh, and Noah~A Smith. 2019.
\newblock Knowledge enhanced contextual word representations.
\newblock \emph{arXiv preprint arXiv:1909.04164}.

\bibitem[{Saxena et~al.(2021)Saxena, Chakrabarti, and
  Talukdar}]{saxena2021cronkgqa}
Apoorv Saxena, Soumen Chakrabarti, and Partha Talukdar. 2021.
\newblock Question answering over temporal knowledge graphs.
\newblock In \emph{Proceedings of the 59th Annual Meeting of the Association
  for Computational Linguistics}.

\bibitem[{Sun et~al.(2020)Sun, Shao, Qiu, Guo, Hu, Huang, and
  Zhang}]{sun2020colake}
Tianxiang Sun, Yunfan Shao, Xipeng Qiu, Qipeng Guo, Yaru Hu, Xuanjing Huang,
  and Zheng Zhang. 2020.
\newblock Colake: Contextualized language and knowledge embedding.
\newblock \emph{arXiv preprint arXiv:2010.00309}.

\bibitem[{Toutanova et~al.(2015)Toutanova, Chen, Pantel, Poon, Choudhury, and
  Gamon}]{toutanova2015representing}
Kristina Toutanova, Danqi Chen, Patrick Pantel, Hoifung Poon, Pallavi
  Choudhury, and Michael Gamon. 2015.
\newblock Representing text for joint embedding of text and knowledge bases.
\newblock In \emph{Proceedings of the 2015 conference on empirical methods in
  natural language processing}, pages 1499--1509.

\bibitem[{Tresp et~al.(2015)Tresp, Esteban, Yang, Baier, and
  Krompa{\ss}}]{tresp2015learning}
Volker Tresp, Crist{\'o}bal Esteban, Yinchong Yang, Stephan Baier, and Denis
  Krompa{\ss}. 2015.
\newblock Learning with memory embeddings.
\newblock \emph{arXiv preprint arXiv:1511.07972}.

\bibitem[{Wang et~al.(2021)Wang, Gao, Zhu, Zhang, Liu, Li, and
  Tang}]{wang2021kepler}
Xiaozhi Wang, Tianyu Gao, Zhaocheng Zhu, Zhengyan Zhang, Zhiyuan Liu, Juanzi
  Li, and Jian Tang. 2021.
\newblock Kepler: A unified model for knowledge embedding and pre-trained
  language representation.
\newblock \emph{Transactions of the Association for Computational Linguistics},
  9:176--194.

\bibitem[{Wu et~al.(2016)Wu, Schuster, Chen, Le, Norouzi, Macherey, Krikun,
  Cao, Gao, Macherey et~al.}]{wu2016google}
Yonghui Wu, Mike Schuster, Zhifeng Chen, Quoc~V Le, Mohammad Norouzi, Wolfgang
  Macherey, Maxim Krikun, Yuan Cao, Qin Gao, Klaus Macherey, et~al. 2016.
\newblock Google's neural machine translation system: Bridging the gap between
  human and machine translation.
\newblock \emph{arXiv preprint arXiv:1609.08144}.

\bibitem[{Xu et~al.(2020)Xu, Nayyeri, Alkhoury, Shariat~Yazdi, and
  Lehmann}]{xu-etal-2020-tero}
Chengjin Xu, Mojtaba Nayyeri, Fouad Alkhoury, Hamed Shariat~Yazdi, and Jens
  Lehmann. 2020.
\newblock \href {https://doi.org/10.18653/v1/2020.coling-main.139} {{T}e{R}o: A
  time-aware knowledge graph embedding via temporal rotation}.
\newblock In \emph{Proceedings of the 28th International Conference on
  Computational Linguistics}, pages 1583--1593, Barcelona, Spain (Online).
  International Committee on Computational Linguistics.

\bibitem[{Xu et~al.(2019)Xu, Nayyeri, Alkhoury, Yazdi, and
  Lehmann}]{xu2019temporal}
Chengjin Xu, Mojtaba Nayyeri, Fouad Alkhoury, Hamed~Shariat Yazdi, and Jens
  Lehmann. 2019.
\newblock Temporal knowledge graph embedding model based on additive time
  series decomposition.
\newblock \emph{arXiv preprint arXiv:1911.07893}.

\bibitem[{Yang et~al.(2014)Yang, Yih, He, Gao, and Deng}]{yang2014embedding}
Bishan Yang, Wen-tau Yih, Xiaodong He, Jianfeng Gao, and Li~Deng. 2014.
\newblock Embedding entities and relations for learning and inference in
  knowledge bases.
\newblock \emph{arXiv preprint arXiv:1412.6575}.

\bibitem[{Yao et~al.(2019)Yao, Mao, and Luo}]{yao2019kg}
Liang Yao, Chengsheng Mao, and Yuan Luo. 2019.
\newblock Kg-bert: Bert for knowledge graph completion.
\newblock \emph{arXiv preprint arXiv:1909.03193}.

\bibitem[{Zhang et~al.(2019)Zhang, Han, Liu, Jiang, Sun, and
  Liu}]{zhang2019ernie}
Zhengyan Zhang, Xu~Han, Zhiyuan Liu, Xin Jiang, Maosong Sun, and Qun Liu. 2019.
\newblock Ernie: Enhanced language representation with informative entities.
\newblock \emph{arXiv preprint arXiv:1905.07129}.

\end{thebibliography}
\bibliographystyle{acl_natbib}

\appendix
\newpage
\section*{\textbf{Appendix}}

\section{Related Work of Temporal Knowledge Embedding}
\label{app: TKE}
Temporal Knowledge Embedding (tKE) is also termed Temporal Knowledge Representation Learning (TKRL), which is to embed entities and predicates of temporal knowledge graphs into low-dimensional vector spaces. TKRL is an expressive and popular paradigm underlying many KG models. To capture temporal aspects, each model either embeds discrete timestamps into a vector space or learns time-dependent representations for each entity. \citet{ma2019embedding} developed extensions of static knowledge graph models by adding timestamp embeddings to their score functions. Besides,  HyTE \citep{dasgupta2018hyte} embeds time information in the entity-relation space by learning a temporal hyperplane to each timestamp and projects the embeddings of entities and relations onto timestamp-specific hyperplanes. 
Later, \citet{goel2020diachronic}  equipped static models with a diachronic entity embedding function which provides the characteristics of entities at any point in time and achieves strong results. Moreover, \citet{han2020dyernie} introduced a non-Euclidean embedding approach that learns evolving entity representations in a product of Riemannian manifolds. It is the first work to contribute to geometric embedding for tKG and achieves state-of-the-art performances on the benchmark datasets. In particular, ECOLA is model-agnostic, which means any temporal KG embedding model can be potentially enhanced by training with the knowledge-text task.
      
\section{ECOLA-SF: an ablation study on static fusion}
\label{app: ecola-sf}
We compare the effectiveness of enhancing temporal knowledge embedding and enhancing static knowledge embedding. In particular, we only feed the static part of entity embeddings into PLM to perform the knowledge-text prediction task. We refer to it as ECOLA-SF (\textbf{S}tatic\textbf{F}usion).

\textbf{ECOLA-SF} is the static counterpart of ECOLA-DE, where we do not apply temporal knowledge embedding to the knowledge-text prediction objective $\mathcal L_{KTP}$. Specifically, we randomly initialize an embedding vector $\mathbf{\bar{e}}_i \in \mathbb R^d$ for each entity $e_i \in \mathcal E$, where $\mathbf{\bar{e}}_i$ has the same dimension as the token embedding in pre-trained language models. Then we learn the \textbf{time-invariant part} $\mathbf{\bar{e}}_i$ via the knowledge-text prediction task. For the tKE objective, we have the following temporal knowledge embedding, 
\begin{equation*}
\mathbf e_i^{SF}(t)[n] = \left\{  
           \begin{aligned}
            &\mathbf W_{sf}{\bar{\mathbf e}}_i[n]  \;\;\;\; \text{if} \;\; 1 \le n \le \gamma d, \\
		  &\mathbf a_{e_i}[n] \sin (\boldsymbol \omega_{e_i}[n]t + \mathbf b_{e_i}[n]) \; \text{else,} \\
\end{aligned}
\right.   
\end{equation*}
where $\mathbf e_i^{SF}(t) \in \mathbb R^d$ is an entity embedding containing static and temporal embedding parts. $\mathbf a_{e_i}, \boldsymbol \omega_{e_i}, \mathbf b_{e_i} \in \mathbb R^{d - \gamma d}$ are entity-specific vectors with learnable parameters. $\mathbf W_{sf} \in \mathbb R^{d\times \gamma d}$ is matrix with learnable weights. Note that $\mathbf e_i^{SF}(t)$ only plays a role in $\mathcal L_{tKE}$, and we use static embedding $\mathbf{\bar{e}}_i$ instead of $\mathbf e_i^{SF}(t)$ in $\mathcal L_{KTP}$. 

\section{Implementation}
\label{app: experimental setting}
We use the datasets augmented with reciprocal relations to train all baseline models. We tune the hyperparameters of our models using the random search and report the best configuration. Specifically,  we set the loss weight $\lambda$ to be 0.3, except for ECOLA-DE model trained on Wiki dataset where $\lambda$ is set to be 0.001. We use the Adam optimizer \citep{kingma2014adam}.
We use the implementation of DE-SimplE\footnote{https://github.com/BorealisAI/de-simple}, ATiSE/TeRO\footnote{https://github.com/soledad921/ATISE}. We use the code for TNTCopmlEx from the tKG framework \citep{han-etal-2021-time}. We implement TTransE based on the implementation of TransE in PyKEEN\footnote{https://github.com/pykeen/pykeen}. 
We tune the model across a range of hyperparameters
as shown in Table \ref{tab:Search space of hyperparameters}. 
We provide the detailed settings of hyperparameters of each baseline model and ECOLA in Table \ref{tab:Hyperparameter configurations of static baselines} in the appendix.  


\begin{table}[htbp]
    \caption{The runtime of the training procedure (in hours).}
    \label{tab: runtime}
     \begin{center}
      \resizebox{.7\columnwidth}{!}{
    \begin{tabular}{lccc} 
      \toprule 
       \multicolumn{1}{l}{Dataset} & \multicolumn{1}{c}{\textbf{GDELT}} & \multicolumn{1}{c}{\textbf{DuEE}} & \multicolumn{1}{c}{\textbf{Wiki}}\\
       \midrule
      \midrule 
       DE-SimplE & 17 & 0.5 & 5.0 \\
       ECOLA-DE & 24.0 & 16.7 &  43.2 \\
       UTEE & 67.3 & 0.5 & 11.3 \\ 
       ECOLA-UTEE & 36.0 & 12.8 & 45.6  \\
       DyERNIE & 25 & 0.1 & 5.9 \\
       ECOLA-DyERNIE & 23.8 & 10.8 & 67.2 \\
      \bottomrule 
    \end{tabular}}
     \end{center}
\end{table}

\begin{table}[]
\centering
\caption{Search space of hyperparameters}
\label{tab:Search space of hyperparameters}
\begin{tabular}{ll}
\hline
Hyperparameter              & Search space                  \\ \hline
learning rate               & \{e-5, 5e-5, e-4, 5e-4, e-3\} \\
warm up                     & \{0.05, 0.2, 0.3\}            \\
weight decay                & \{0.01, 0.05, 0.2\}           \\
batch size                  & \{16, 128, 256, 512\}         \\ \hline
\end{tabular}
\end{table} 

\section{The amount of Compute and the Type of Resources Used}
We run our experiments on an NVIDIA A40 with a memory size of 48G. We provide the \textbf{training time} of our models and some baselines in Table \ref{tab: runtime}. Note that there are no textual descriptions at inference time, and we take the entity and predicate embedding as input and use the score function of KG models to predict the missing links. Thus, the inference time of ECOLA (e.g., ECOLA-DE) and its counterpart KG model (e.g., DE-SimplE) is the \textbf{same}.
The numbers of parameters are in Table \ref{tab:number of parameters} .
\begin{table}[]
\centering
\caption{The number of parameters ($M$).}
\label{tab:number of parameters}
\begin{tabular}{llll}
\hline
Dataset & GDELT & DuEE & Wiki \\ \hline
DyERNIE & 159   & 140  & 174  \\
UTEE    & 150   & 139  & 158  \\
DE      & 175   & 140  & 173  \\ \hline
\end{tabular}
\end{table}

\begin{table*}[htbp]
  \begin{center}
    \caption{Datasets Statistics }
    \label{tab:dataset statistics}
    \resizebox{140mm}{10mm}{
    \begin{tabular}{ccccccccc}
      \toprule 
      Dataset & \# Entities  & \# Predicates & \# Timestamps  & \# training set & \# validation set & \# test set \\
      \midrule 
      GDELT & 5849 & 237 & 2403 & 755166 & 94395 & 94395  \\
      DUEE & 219 & 41 & 629 & 1879 & 247 & 247 \\
      WIKI & 10844 & 23 & 82 & 233525 & 19374 & 19374 \\
      \bottomrule 
    \end{tabular}
}
  \end{center}
 \end{table*}

\begin{table*}[htbp]
 \caption{Hyperparameter settings of ECOLA and baselines.}
    \label{tab:Hyperparameter configurations of static baselines}
\resizebox{\linewidth}{!}{
\begin{tabular}{l|ccc|ccc|ccc|ccc} 
    \toprule
    \multicolumn{1}{l}{Parameters}&\multicolumn{3}{|c}{Embedding dimension}&\multicolumn{3}{|c}{Negative Sampling}&\multicolumn{3}{|c}{Learning rate}&\multicolumn{3}{|c}{Batch Size}\\
    \midrule
    Datasets & GDELT & DuEE & Wiki & GDELT & DuEE & Wiki & GDELT & DuEE & Wiki & GDELT & DuEE & Wiki\\
    \midrule
    TransE & 768 & 768 & 768 & 200 & 100 & 100 & 5e-4 & 5e-4 & 5e-4 &256 & 128 & 256 \\
    SimplE & 768 & 768 & 768 & 200 & 100 & 100 & 5e-4 & 5e-4 & 5e-4 &256 & 128 & 256 \\
    TTransE & 768 & 768 & 768 & 200 & 100 & 100 & 5.2e-4 & 5.2e-4 & 5.2e-4 & 256 & 256 & 256 \\
    TNTComplEx & 768 & 768 & 768 & 200 & 100 & 100 & 1.5e-4 & 1.5e-4 & 1.5e-4 & 256 & 256 & 256 \\
    DE-SimplE & 768 & 768 & 768 & 200 & 100 & 100 & 5e-4 & 5e-4 & 5e-4 &256 & 128 & 256 \\
    ECOLA-SF & 768 & 768 & 768 & 200 & 100 & 100 & 1e-4 & 2e-5 & 1e-4 & 64 & 16 & 64 \\
    ECOLA-DE & 768 & 768 & 768 & 200 & 200 & 200 & 2e-5 & 2e-5 & 2e-5 & 4 & 8 & 4 \\ 
    ECOLA-UTEE & 768 & 768 & 768 & 200 & 200 & 200 & 2e-5 & 2e-5 & 2e-5 & 4 & 8 & 4 \\ 
    ECOLA-dyERNIE & 768 & 768 & 768 & 200 & 200 & 200 & 2e-5 & e-4 & 2e-5 & 4 & 8 & 4 \\
    \bottomrule
\end{tabular}
}
\end{table*}

\section{The license of the Assets}
We adapt three existing datasets, i.e., GDELT, DuEE, and Wiki. We would first state the original license. 
\begin{itemize}
    \item \textbf{GDELT:} as stated in the term of use of GDELT\footnote{https://www.gdeltproject.org/about.html\#termsofuse}, the GDELT Project is an open platform for research and analysis of global society and thus all datasets released by the GDELT Project are available for unlimited and unrestricted use for any academic, commercial, or governmental use of any kind without fee. One may redistribute, rehost, republish, and mirror any of the GDELT datasets in any form. However, any use or redistribution of the data must include a citation to the GDELT Project and a link to this website (https://www.gdeltproject.org/).
    
    \item \textbf{Wiki} is proposed by \citet{dasgupta2018hyte} and has Apache License 2.0.
    
    \item \textbf{DuEE} is released by Baidu Research. As stated on its website\footnote{https://ai.baidu.com/broad/introduction?dataset=duee}, they have committed to provide these datasets at no cost for research and personal uses.
\end{itemize}

For the derived datasets, we only release a short version due to the size limit of uploads. Thus, we will release the full version and give the license, copyright information, and terms of use once the paper gets accepted. 

\section{Documentation of the artifacts}
This paper uses three datasets, GDELT, Wiki, and DuEE. GDELT mainly covers social and political events written in English. Wiki in this paper mainly contains evolving knowledge, i.e., affiliation and residence place information, which is also written in English. DuEE is a dataset in Chinese and mainly talks about social news, such as the launch of new electronic products.

\end{document}